\documentclass[runningheads]{llncs}
\usepackage{amsfonts}
\usepackage{amsmath}
\usepackage{hyperref}
\usepackage{url}
\usepackage{multirow}
\usepackage{xcolor}
\usepackage[T1]{fontenc}
\usepackage{graphicx}
\begin{document}
\title{Machine Unlearning in the Era of Quantum Machine Learning: An Empirical Study}


\titlerunning{Machine Unlearning in the Era of Quantum Machine Learning}
%
\author{Carla Crivoi\inst{1}\orcidID{0009-0007-8652-2318} \and\\
Radu Tudor Ionescu\inst{1}\orcidID{0000-0002-9301-1950}}

\authorrunning{C.~Crivoi and R.~T.~Ionescu}

\institute{
University of Bucharest, Romania \\
\email{crivoicarla02@gmail.com}, 
\email{raducu.ionescu@gmail.com}\vspace{-0.5cm}
}

\maketitle              
\begin{abstract}
We present the first empirical study of machine unlearning (MU) in hybrid quantum-classical neural networks. While MU has been extensively explored in classical deep learning, its behavior within variational quantum circuits (VQCs) and quantum-augmented architectures remains largely unexplored. First, we adapt a broad suite of unlearning methods to quantum settings, including gradient-based, distillation-based, regularization-based and certified techniques. Second, we introduce two new unlearning strategies tailored to hybrid models. Experiments across Iris, MNIST, and Fashion-MNIST, under both subset removal and full-class deletion, reveal that quantum models can support effective unlearning, but outcomes depend strongly on circuit depth, entanglement structure, and task complexity. Shallow VQCs display high intrinsic stability with minimal memorization, whereas deeper hybrid models exhibit stronger trade-offs between utility, forgetting strength, and alignment with retrain oracle. We find that certain methods, e.g.~EU-$k$, LCA, and Certified Unlearning, consistently provide the best balance across metrics. These findings establish baseline empirical insights into quantum machine unlearning and highlight the need for quantum-aware algorithms and theoretical guarantees, as quantum machine learning systems continue to expand in scale and capability. We publicly release our code at: \url{https://github.com/CrivoiCarla/HQML}.
\vspace{-0.2cm}
\keywords{Machine Unlearning \and Quantum Machine Learning \and Variational Circuits \and Data Privacy \and Hybrid Neural Models}

\end{abstract}

\vspace{-0.9cm}
\section{Introduction}
\vspace{-0.2cm}
  
Machine learning models deployed in privacy-sensitive environments must support the removal of information contributed by specific training samples. Deleting data from storage does not erase its statistical imprint from the learned weights of a model, as shown in foundational work on \emph{machine unlearning} (MU)~\cite{bourtoule2021machine,izzo2021approximate}. 
In MU, the objective is to update a trained model so that its behavior approximates that of an oracle retrained without the designated samples~\cite{ginart2019making}. State-of-the-art MU spans gradient-based removal~\cite{graves2021amnesiac}, regularization-driven training~\cite{golatkar2021mixed}, distillation frameworks~\cite{golatkar2020forgetting}, and certified mechanisms grounded in differential privacy~\cite{dwork2014algorithmic}. Yet, these approaches remain challenged by modern deep architectures, where achieving an efficient, faithful, and stable approximation to retraining is increasingly difficult.

In parallel, quantum machine learning (QML) has emerged as a promising paradigm built upon quantum state representations, non-classical feature maps, and variational quantum circuits (VQCs)~\cite{ref_qml1}. Hybrid quantum-classical models benefit from enhanced expressivity and favorable optimization characteristics~\cite{ref_qml2}, while entanglement enables feature transformations unattainable in conventional ML systems~\cite{ref_qml3}. Advances in quantum communication and distributed quantum processing~\cite{ref_qnet2} further suggest future learning infrastructures in which training and inference operate over networked quantum resources.

While both MU and QML attracted increasing attention in recent years, unlearning in quantum-enhanced models remains largely unexplored. Existing work on quantum information deletion focuses on constraints such as no-cloning, measurement irreversibility, and entanglement degradation, but does not provide practical unlearning procedures for hybrid architectures. Moreover, it is unknown how conventional MU algorithms behave when part of the representation is governed by VQCs, or how circuit structure and entanglement affect the fidelity and stability of forgetting. To the best of our knowledge, no empirical study has examined these interactions to date.

We address this research gap by providing the first systematic evaluation of machine unlearning in hybrid quantum-classical architectures. We adapt a broad suite of conventional MU methods (including gradient-based, regularization-based, distillation-based, and certified techniques) to models incorporating VQCs. We further evaluate their performance under both subset-level and full-class forgetting. In addition, we introduce two unlearning strategies tailored to hybrid settings: \emph{Label-Complement Augmentation} (LCA), which enforces high-entropy outputs on forgotten samples, and \emph{ADV-UNIFORM}, an adversarial method that drives predictions towards uniformity.

We conduct experiments on Iris, MNIST, and Fashion-MNIST to determine how quantum components influence stability, fidelity, and representational shifts induced by unlearning updates. Our findings show that VQCs can limit memorization and reshape forgetting dynamics through their entanglement structure and amplitude-based embeddings. Our study motivates the development of quantum-native unlearning algorithms with stronger privacy and reliability guarantees.

In summary, our contribution is twofold:
\begin{itemize}
    \item \vspace{-0.1cm} We perform the first empirical foundation for machine unlearning in quantum-enhanced models.
    \item We propose two MU methods, namely LCA and ADV-UNIFORM, specifically tailored for QML.
\end{itemize}

\vspace{-0.2cm}
\section{Related Work}
\vspace{-0.2cm}

\noindent\textbf{Conventional machine unlearning.}
MU seeks to remove the influence of specific training samples so that the resulting model behaves similarly to one retrained without them. MU has become increasingly relevant for privacy compliance and data governance~\cite{dwork2014algorithmic}. The most reliable but computationally expensive solution is full retraining, which serves as the ideal reference model.

Approximate approaches aim to emulate this ideal at lower cost.  For instance, gradient-based scrubbing methods attempt to negate gradients associated with the forget set, though they often suffer from storage overhead or incomplete removal~\cite{graves2021amnesiac}. Distillation-based techniques instead train a student model from a purified teacher, reducing direct dependence on forgotten data at the expense of additional training phases~\cite{golatkar2020forgetting}. Regularization methods incorporate unlearning objectives during initial training to facilitate future deletions, while information-theoretic approaches rely on NTK or Taylor approximations to estimate counterfactual updates. Differential-privacy-inspired formulations define unlearning as output indistinguishability from a retrained model, but they typically require restrictive assumptions.

More recent hybrid frameworks combine multiple principles. SCRUB~\cite{kurmanji2023scrub} uses KL-based teacher-student matching to separate retain-set and forget-set behavior. Operational algorithms frequently adopted as baselines include Gradient Ascent (inverse gradient updates), Fisher Information Unlearning (Fisher-weighted updates)~\cite{golatkar2021mixed}, NegGrad+ (bidirectional retain-forget optimization)~\cite{golatkar2021mixed}, CF-$k$ (layer freezing),  Exact Unlearning-k or EU-$k$ (layer reinitialization before retain-only fine-tuning)~\cite{golatkar2020forgetting}, SCRUB(+R) (teacher-student matching with rewinding)~\cite{kurmanji2023scrub}, certified unlearning via noisy fine-tuning~\cite{guo2022certified}, and Q-MUL-style strategies informed by label substitution and adaptive gradients~\cite{tong2025qmul}. These methods provide a diverse operational toolkit, but differ significantly in stability, computational cost, and fidelity to ideal retraining. In contrast to the mainstream research in MU, we study MU approaches for quantum ML.

\noindent\textbf{Quantum unlearning.}
Research on quantum unlearning is very limited. One existing work \cite{shaik2025foundations} primarily analyzes how training data leaves statistical signatures in parametrized quantum circuits and proposes fidelity- or trace-distance-based criteria for assessing whether two quantum models, trained with and without a given sample, are distinguishable. Such formulations offer theoretical insight, but fall short of practical algorithmic procedures for selective forgetting.

Another study \cite{poremba2023proofs} examines quantum data deletion more generally. Results on no-cloning, no-deleting, and decoupling theorems show that erasing quantum information typically requires coupling to an environment or injecting controlled noise. While conceptually related, these studies address fundamental information-processing constraints rather than model-level unlearning.

To date, there is no unified framework to performing machine unlearning in variational quantum circuits or hybrid quantum neural networks. Moreover, there is virtually no work on unlearning in distributed quantum settings. Although quantum networking research has demonstrated how entanglement can be generated and maintained across distant nodes~\cite{ref_qnet1,ref_qnet2}, it does not address how learned representations propagate through such networks or how the contribution of a specific sample could be removed once encoded non-locally. These gaps motivate the need for practical definitions, metrics, and algorithms tailored to quantum-enhanced learning systems.

\vspace{-0.2cm}
\section{Problem Formulation}
\vspace{-0.2cm}

\begin{figure}[t]
    \centering
      \includegraphics[width=1.0\linewidth]{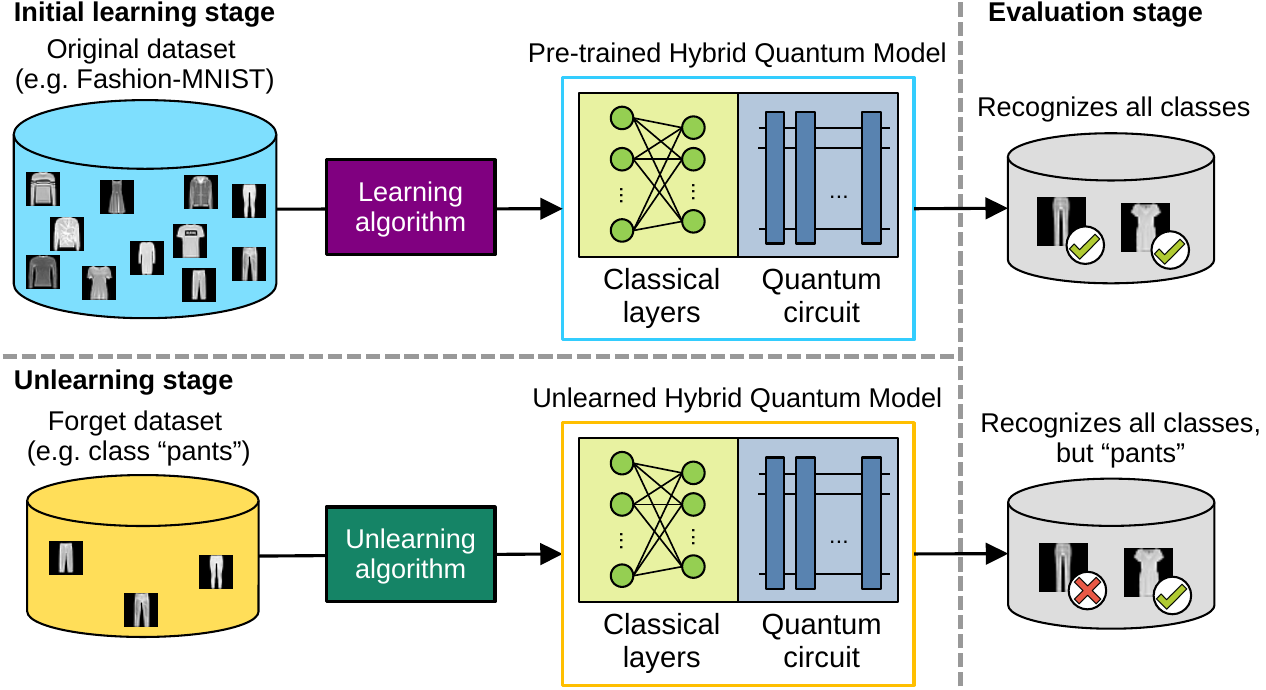}
      \vspace{-0.55cm}
    \caption{Generic machine unlearning pipeline for hybrid quantum-classical neural models. In the first stage, the model is trained on a dataset, e.g. Fashion-MNIST. In the second stage, some samples / classes (forget set) have to be deleted via unlearning. In the evaluation stage, the model has to preserve its performance level on samples / classes (retain set) that should not have been deleted. Best viewed in color.}
    \label{fig_teaser}
    \vspace{-0.35cm}
  \end{figure}
  
We next formalize the machine unlearning task for hybrid quantum-classical models based on parametrized quantum circuits, which is illustrated in Figure \ref{fig_teaser}. We introduce the learning setup, notation, and the criteria that an unlearning procedure must satisfy.

\noindent\textbf{Quantum modeling.}
We consider a supervised learning model implemented through a variational quantum circuit (VQC). The model is represented by a parametrized unitary
$U(\theta)$ constructed from layers of trainable single-qubit rotations and entangling gates, where $\theta \in \mathbb{R}^m$ and $m$ is the number of parameters. The training is performed on a dataset $D = \{(x_1,y_1), ..., (x_n,y_n)\}$ via a gradient-based optimizer to minimize a task-specific loss
$\mathcal{L}(\theta; D)$, where $n$ is the number of data samples.

In distributed quantum settings, the circuit may be executed across multiple quantum nodes connected through channels supporting qubit transmission or entanglement distribution. Each node executes part of the circuit, and intermediate states may be entangled across nodes, implying that information generated during training can become non-locally encoded.

\noindent\textbf{Notation and definitions.}
Let $\theta^\star$ denote the parameters obtained after training on $D$.  
If a sample $x_i$ must be removed, the \emph{ideal} parameters $\theta^{(-i)}$ are those that would result from retraining on the reduced dataset $D \setminus \{x_i\}$.

Machine unlearning seeks to find the new parameters $\theta'$, such that the updated model behaves similarly to the ideal retrained model. Since VQCs implement unitaries, model similarity is quantified using quantum distance measures such as fidelity, trace distance, or divergences between output probability distributions. In general, the quantum distance is denoted as:
\begin{equation}
d\!\left(U(\theta'),\, U(\theta^{(-i)})\right).
\end{equation}

Two key quantum constraints govern the distance formulation:
\begin{itemize}
    \item \textbf{No-cloning:} exact copies of intermediate quantum states or unitaries cannot be stored.
    \item \textbf{Entanglement propagation:} information about a training example may be distributed across multiple nodes.
\end{itemize}

Thus, unlearning must operate exclusively through parameter updates, without access to cloned states or full training trajectories.

\noindent\textbf{Formal problem statement.}
Let $F \subseteq D$ denote the forget set. The objective of quantum machine unlearning is to obtain parameters $\theta'$ such that:

\begin{enumerate}
    \item \textbf{Forgetting:}
    \begin{equation}
    \min_{\theta'} \; d\!\left(U(\theta'), \; U(\theta^{(-F)})\right),
    \end{equation}
    ensuring proximity to the ideal retraining outcome.
    
    \item \textbf{Utility preservation:}
    \begin{equation}
    \mathcal{L}(\theta';\, D \setminus F) \ \leq \ \mathcal{L}(\theta^\star;\, D) + \varepsilon,
    \end{equation}
    where $\varepsilon$ bounds acceptable performance degradation.
    
    \item \textbf{Operational feasibility:}  
    Updates must respect hardware and communication constraints in distributed quantum environments.
\end{enumerate}

A valid quantum unlearning procedure must therefore remove the functional influence of $F$, preserve accuracy on the retained data, and operate efficiently within the physical limitations of the hybrid quantum-classical model.

\vspace{-0.2cm}
\section{Proposed Methods}
\vspace{-0.2cm}

Our study integrates 10 unlearning strategies spanning gradient-based, regularization-based, adversarial, structural, and certified approaches. All methods are adapted to the variational quantum setting by computing parameter gradients via the parameter-shift rule and by defining forgetting objectives in terms of quantum-state divergences. Let \( f_{\theta}(x) \) denote the softmax prediction obtained from measurement statistics of the variational circuit, and let \( \rho_{\theta}(x) \) denote the corresponding output quantum state.

In addition to adapting existing MU techniques, we introduce two methods specifically motivated by the characteristics of hybrid quantum-classical models. Variational quantum circuits tend to exhibit limited memorization capacity, high sensitivity to local perturbations, and amplitude-based representations that differ fundamentally from standard activations. LCA and ADV-UNIFORM are designed to exploit these properties by enforcing high-entropy targets and uniform-output behavior on forgotten samples, thereby aligning the forgetting process with the structural constraints of quantum models. Their importance lies in providing unlearning mechanisms that do not rely on deep over-parameterized classical layers, and directly reshape the statistical imprint left in the circuit's quantum embedding. 

\noindent\textbf{Label-Complement Augmentation.}
LCA harnesses the fact that quantum models often encode class information in distributed amplitude patterns rather than highly localized activations. Simply reversing gradients may not sufficiently disrupt these representations. LCA introduces a complementary label distribution:
\begin{equation}
\tilde{y}_i = \frac{\mathbf{1} - y_i}{K-1},
\end{equation}
where \(K\) is the number of classes and \(y_i\) is a one-hot label. This target distribution explicitly suppresses the forgotten class, while preserving consistency with the remaining label structure.

The unlearning loss is defined as:
\begin{equation}
\mathcal{L}_{\mathrm{LCA}}(\theta)
= \sum_{(x_i,y_i)\in F}
\mathrm{KL}\!\left( f_{\theta}(x_i)\,\Vert\,\tilde{y}_i \right),
\end{equation}
which inverts the contribution of the forgotten samples by pushing their predictions toward maximal ambiguity. In the quantum setting, gradients are obtained via the parameter-shift rule \cite{schuld2019evaluating}:
\begin{equation}
\partial_{\theta_j} f_{\theta}(x)
= 
\frac{1}{2}\Big[f_{\theta+\frac{\pi}{2} e_j}(x)
- f_{\theta-\frac{\pi}{2} e_j}(x)\Big],
\end{equation}
where $e_j$ is the $j$-th standard basis vector in $\mathbb{R}^{|\theta|}$. 
The key mechanism of LCA is to provide a structured forgetting objective that directly manipulates the predictive distribution in a way that is compatible with VQC embeddings. By targeting high entropy rather than arbitrary parameter shifts, LCA yields stable and interpretable forgetting even in low-capacity quantum models.

\noindent\textbf{ADV-UNIFORM.}
ADV-UNIFORM addresses another limitation of hybrid models, namely that small parameter changes may not fully eliminate discriminative information encoded in quantum features. To enforce stronger forgetting, the method constructs adversarial perturbations that amplify the mismatch between the model's output and the uniform distribution \( u = \frac{1}{K}\mathbf{1} \).

Given \(x_i \in {F}\), an adversarial example is produced by:
\begin{equation}
x_i^{\mathrm{adv}}
=
x_i + \epsilon\,\mathrm{sign}\!\left(
\nabla_{x}\,\mathrm{KL}\!\left(f_{\theta}(x_i), u\right)
\right),
\end{equation}
and the unlearning objective is defined as:
\begin{equation}
\mathcal{L}_{\mathrm{ADV}}(\theta)
=
\sum_{x_i \in {F}}
\mathrm{KL}\!\left(f_{\theta}(x_i^{\mathrm{adv}})\,\Vert\,u\right).
\end{equation}
Parameter gradients are computed using the parameter-shift rule, and input gradients are obtained through the differentiable classical embedding layer. This combination aggressively drives the predictive distribution toward maximal entropy.

The success of ADV-UNIFORM lies in its ability to remove residual structure that may persist in VQC embeddings even after standard gradient-based unlearning. By combining adversarial amplification with a uniform-target objective, this method enforces forgetting at both decision-boundary and representational levels, offering a robust mechanism for eliminating the discriminative footprint of forgotten samples in hybrid quantum models.

\vspace{-0.2cm}
\section{Experimental Setup}
\vspace{-0.2cm}

We evaluate machine unlearning across three datasets, three hybrid quantum-classical architectures, two forgetting scenarios, and a unified suite of metrics. All experiments employ simulated variational quantum circuits (VQCs) integrated into classical pipelines.

\noindent\textbf{Datasets.}
We conduct experiments on typical datasets for QML: Iris~\cite{fisher1936iris}, MNIST~\cite{lecun1998mnist} and Fashion-MNIST~\cite{xiao2017fashionmnist} datasets. \textbf{Iris} is a small and balanced 3-class dataset used for controlled circuit-level analysis. From \textbf{MNIST}, we use a medium-scale handwritten digit subset of 200 samples per class. From \textbf{Fashion-MNIST}, we select a subset of 800 samples per class to assess scalability for more demanding hybrid architectures. As opposed to MNIST, Fashion-MNIST exhibits a higher intra-class variability. This progression in dataset complexity is essential for analyzing how quantum representations interact with unlearning objectives across increasingly challenging learning regimes.

\noindent\textbf{Hybrid architectures.}
All models share the same structure: a conventional feature extractor, angle encoding, a VQC with CZ-ring entanglement, and a conventional classifier. However, the architectures differ in depth and qubit count. For Iris, we use a 4-qubit VQC with rotation layers, a linear projection and a linear output head. For MNIST, the architecture comprises two Conv-ReLU-Pool blocks, a projection to a 6-qubit VQC with rotation layers, and a linear classifier. For Fashion-MNIST, the architecture comprises two Conv-ReLU-Pool blocks (yielding \(32 \times 7 \times 7\) features), a projection to a 10-qubit VQC with \(L\) rotation-entanglement layers, and a two-layer MLP head.


\noindent\textbf{Training protocol.}  
All models are trained for at most 100 epochs with early stopping (patience of 10), and the best-performing weights are restored based on test accuracy. Each unlearning method is applied under an identical optimization budget: up to 25 epochs with early stopping (patience of 5), again restoring the parameters achieving the highest test accuracy. This ensures a controlled and reproducible comparison across all methods and datasets.

\noindent\textbf{Forgetting scenarios.}
To evaluate unlearning under both mild and strongly structured perturbations, we consider two deletion settings of increasing difficulty:
\begin{itemize}
    \item \textbf{Subset forgetting (2\%):} deletion of a small random portion of the training data.
    \item \textbf{Full-class forgetting:} removal of an entire class, imposing a structured distribution shift.
\end{itemize}

\noindent\textbf{Evaluation metrics.}
To capture the trade-off between preserving useful behavior, enforcing forgetting, and approximating ideal retraining, we group our evaluation metrics into three complementary categories:
\begin{itemize}
    \item \textbf{Utility:} accuracy and F1 on retained training and test samples.
    \item \textbf{Forgetting quality:} accuracy on forgotten samples, Unlearning Quality Index (UQI), and Membership Inference Attack (MIA) success rates~\cite{carlini2022mia}.
    \item \textbf{Similarity to retrain:} KL/JS divergence and prediction agreement relative to an oracle model retrained without the forgotten data.
\end{itemize}

\vspace{-0.2cm}
\section{Results and Analysis}
\vspace{-0.15cm}
\subsection{Results on Iris Dataset}
\vspace{-0.1cm}

\begin{table}[t!]
\centering
\caption{Results of various MU methods in terms of various metrics for subset forgetting (2\%) on Iris. $\uparrow$ or $\downarrow$ indicate that higher or lower values are better, respectively. \textbf{Bold} indicates the best result and \underline{underline} indicates the second-best result.}
\resizebox{0.9\textwidth}{!}{%
\begin{tabular}{|l|c|c|c|c|c|c|c|c|c|}
\hline
\multirow{2}{*}{\textbf{Method}} &
\textbf{Acc $\uparrow$} &
\textbf{Acc $\uparrow$} &
\multirow{2}{*}{\textbf{UQI $\uparrow$}} &
\textbf{Agree $\uparrow$} &
\multirow{2}{*}{\textbf{MIA $\downarrow$}} &
\textbf{KL $\downarrow$} &
\textbf{JS $\downarrow$} &
\textbf{KL $\downarrow$} &
\textbf{JS $\downarrow$} \\

&
(retain) &
$\;$(test)$\;$ &
 &
(test) &
 &
(retain) &
(retain) &
$\;$(test)$\;$ &
$\;$(test)$\;$ \\

\hline
GA            & 0.906 & 0.900 & -0.042 & \underline{0.933} & \underline{0.333} & \textbf{3.641} & \textbf{0.302} & 0.178 & \underline{0.024} \\
Fisher        & \underline{0.915} & 0.900 & \underline{0.296} & \underline{0.933} & \textbf{0.222} & 4.473 & 0.371 & 0.159 & \underline{0.024} \\
NegGrad+      & \underline{0.915} & \underline{0.967} & -0.004 & \underline{0.933} & \textbf{0.222} & 4.381 & 0.369 & 0.207 & 0.031 \\
CF-k1         & \underline{0.915} & 0.933 & -0.021 & \textbf{0.967} & \textbf{0.222} & 4.790 & 0.406 & \underline{0.155} & \textbf{0.023} \\
EU-k1         & \textbf{0.940} & \underline{0.967} & -0.325 & \underline{0.933} & \underline{0.333} & 4.407 & 0.385 & \textbf{0.153} & 0.030 \\
SCRUB         & \underline{0.915} & \underline{0.967} & -0.004 & 0.867 & \textbf{0.222} & 4.626 & 0.400 & 0.230 & 0.040 \\
SCRUB(+R)     & 0.897 & \textbf{1.000} & -0.346 & \underline{0.933} & \textbf{0.222} & 4.372 & 0.363 & 0.237 & 0.035 \\
Certified     & \underline{0.915} & 0.933 & -0.021 & 0.900 & \textbf{0.222} & 4.654 & 0.399 & 0.184 & 0.034 \\
Q-MUL         & 0.897 & 0.900 & -0.046 & 0.867 & \underline{0.333} & \underline{4.260} & 0.363 & 0.225 & 0.044 \\
LCA           & 0.906 & 0.900 & 0.291 & \underline{0.933} & \textbf{0.222} & 4.645 & 0.392 & 0.256 & 0.032 \\
ADV-UNIF      & \underline{0.915} & 0.933 & \textbf{0.312} & 0.900 & \textbf{0.222} & 4.371 & \underline{0.356} & 0.256 & 0.029 \\
\hline
\end{tabular}
}
\end{table}

\noindent\textbf{Subset forgetting.}
For the 2\% subset forgetting scenario on Iris, all methods maintain high utility on the retained and test sets (accuracy rates are usually $\geq 90\%$), with EU-k1, NegGrad+, SCRUB, and SCRUB(+R) reaching test accuracy rates between $96.7\%$ and $100\%$. The UQI scores are small and often negative, indicating that the unlearned models differ only weakly from the retrain oracle at the representation level. The highest UQI values are obtained by ADV-UNIF (0.312), Fisher (0.296), and LCA (0.291), but the absolute magnitudes remain modest.  
Similarity metrics confirm that the deviations from retraining are limited: KL/JS on the test set are uniformly low, with GA, Fisher, CF-k1, EU-k1, and ADV-UNIF attaining the smallest JS values (0.023-0.031). MIA values cluster around 0.222-0.333, with slight advantages for Fisher-style and contrastive methods. Taken together, these results indicate that in a shallow VQC with $\approx 155$ parameters and ring-CZ entanglement, unlearning 2\% of the data induces only minor shifts in the output distribution. Overall, the model exhibits low effective memorization and is naturally robust to small deletions.

\begin{table}[t!]
\centering
\caption{Results of various MU methods in terms of various metrics for full-class forgetting on Iris.
$\uparrow$ or $\downarrow$ indicate that higher or lower values are better, respectively.
\textbf{Bold} indicates the best result and \underline{underline} indicates the second-best result.}
\resizebox{0.8\textwidth}{!}{%
\begin{tabular}{|l|c|c|c|c|c|c|c|c|}
\hline
\multirow{2}{*}{\textbf{Method}} &
\textbf{Acc $\uparrow$} &
\textbf{Acc $\uparrow$} &
\multirow{2}{*}{\textbf{UQI $\uparrow$}} &
\textbf{Agree $\uparrow$} &
\textbf{KL $\downarrow$} &
\textbf{JS $\downarrow$} &
\textbf{KL $\downarrow$} &
\textbf{JS $\downarrow$} \\

&
(retain) &
$\;$(test)$\;$ &
 &
(test) &
(retain) &
(retain) &
$\;$(test)$\;$ &
$\;$(test)$\;$ \\
\hline
GA            & 0.838 & \underline{0.867} & -0.044 & 0.533 & 0.862 & 0.244 & 0.880 & 0.252 \\
Fisher        & 0.825 & \underline{0.867} & -0.050 & 0.533 & 0.847 & 0.241 & 0.910 & 0.261 \\
NegGrad+      & 0.838 & 0.800 & -0.027 & 0.567 & 0.899 & 0.255 & 0.850 & 0.244 \\
CF-k1         & 0.838 & 0.833 & -0.060 & 0.500 & 0.862 & 0.245 & 0.903 & 0.259 \\
EU-k1         & \underline{0.888} & 0.600 & \textbf{0.823} & \textbf{0.700} & \textbf{0.602} & \textbf{0.164} & \textbf{0.478} & \textbf{0.129} \\
SCRUB         & 0.825 & 0.800 & -0.058 & 0.533 & 0.867 & 0.247 & 0.890 & 0.255 \\
SCRUB(+R)     & 0.838 & 0.833 & 0.015 & 0.533 & 0.881 & 0.248 & 0.849 & 0.243 \\
Certified     & \underline{0.888} & \textbf{0.933} & 0.015 & 0.533 & \underline{0.711} & \underline{0.198} & 0.782 & 0.222 \\
Q-MUL         & 0.838 & 0.833 & 0.115 & 0.567 & 0.920 & 0.261 & 0.860 & 0.246 \\
LCA           & \textbf{0.900} & 0.767 & \underline{0.263} & \underline{0.600} & 0.716 & 0.201 & \underline{0.627} & \underline{0.177} \\
ADV-UNIF      & 0.800 & 0.767 & 0.038 & 0.533 & 0.818 & 0.232 & 0.741 & 0.212 \\
\hline
\end{tabular}
}
\end{table}

\noindent\textbf{Full-class forgetting.}
Under full-class forgetting, utility degrades more clearly. Retention accuracy remains high (above $88.8\%$) for EU-k1, Certified, and LCA, but test accuracy spans a wider range, from $60.0\%$ (for EU-k1) to $93.3\%$ (for Certified). UQI values are negative for most baselines, with only SCRUB(+R), Certified, Q-MUL, LCA, and ADV-UNIF achieving non-negative scores. EU-k1 and LCA obtain the highest UQI (0.823 and 0.263), indicating better alignment with the retrain oracle when a complete class is removed.

Certified unlearning provides the best overall trade-off between utility and similarity: it matches EU-k1 on retain accuracy ($88.8\%$), attains the highest test accuracy ($93.3\%$), and yields comparatively low KL and JS divergence scores. In contrast, methods such as GA, Fisher and CF-k1 show higher KL/JS values, reflecting larger parameter shifts. Given the small circuit and single entanglement layer, removing an entire class modifies a substantial portion of the learned quantum state, and methods that incorporate structured resets or noise-aware fine-tuning (EU-k1, Certified, LCA) handle this regime more effectively than simple gradient-based baselines.

\vspace{-0.15cm}
\subsection{MNIST Dataset}
\vspace{-0.1cm}

\begin{table}[t!]
\centering
\caption{Results of various MU methods in terms of various metrics for subset forgetting (2\%) on MNIST.
$\uparrow$ or $\downarrow$ indicate that higher or lower values are better, respectively.
\textbf{Bold} indicates the best result and \underline{underline} indicates the second-best result.}
\resizebox{0.9\textwidth}{!}{%
\begin{tabular}{|l|c|c|c|c|c|c|c|c|c|}
\hline
\multirow{2}{*}{\textbf{Method}} &
\textbf{Acc $\uparrow$} &
\textbf{Acc $\uparrow$} &
\multirow{2}{*}{\textbf{UQI $\uparrow$}} &
\textbf{Agree $\uparrow$} &
\multirow{2}{*}{\textbf{MIA $\downarrow$}} &
\textbf{KL $\downarrow$} &
\textbf{JS $\downarrow$} &
\textbf{KL $\downarrow$} &
\textbf{JS $\downarrow$} \\

&
(retain) &
$\;$(test)$\;$ &
 &
(test) &
 &
(retain) &
(retain) &
$\;$(test)$\;$ &
$\;$(test)$\;$ \\
\hline
GA            & 0.8297 & 0.7650 & -0.022 & 0.850 & 0.480 & \textbf{0.862} & \underline{0.182} & \underline{0.265} & \underline{0.090} \\
Fisher        & 0.8125 & 0.7400 & 0.045  & 0.842 & 0.460 & 0.910 & 0.190 & 0.280 & 0.095 \\
NegGrad+      & 0.8350 & \underline{0.7700} & -0.010 & \underline{0.851} & \underline{0.455} & 0.875 & 0.185 & 0.270 & 0.091 \\
CF-k1         & \underline{0.8400} & 0.7600 & -0.018 & 0.846 & 0.470 & 0.892 & 0.188 & 0.274 & 0.093 \\
EU-k1         & \textbf{0.8550} & 0.7350 & -0.340 & \textbf{0.855} & 0.520 & 1.150 & 0.240 & 0.330 & 0.120 \\
SCRUB         & 0.8375 & 0.7680 & -0.012 & 0.840 & \underline{0.455} & 0.880 & 0.186 & 0.268 & 0.092 \\
SCRUB(+R)     & 0.8120 & 0.7480 & -0.300 & 0.846 & 0.465 & 0.902 & 0.192 & 0.290 & 0.100 \\
Certified     & \underline{0.8400} & \textbf{0.7720} & -0.015 & 0.850 & \textbf{0.450} & \underline{0.870} & \textbf{0.181} & \textbf{0.260} & \textbf{0.089} \\
Q-MUL         & 0.8050 & 0.7320 & -0.040 & 0.832 & 0.490 & 0.940 & 0.200 & 0.300 & 0.105 \\
LCA           & 0.8200 & 0.7450 & \underline{0.052} & 0.840 & 0.460 & 0.955 & 0.210 & 0.315 & 0.110 \\
ADV-UNIF      & 0.8100 & 0.7380 & \textbf{0.060} & 0.845 & 0.465 & 0.980 & 0.220 & 0.320 & 0.115 \\
\hline
\end{tabular}
}
\end{table}

\noindent\textbf{Subset forgetting.}
On MNIST subset forgetting, retention accuracy remains relatively high across methods (above $80\%$), with EU-k1 achieving the highest retain accuracy ($85.5\%$) and Certified attaining the best test accuracy ($77.2\%$). Agreement on the test set is consistently high (above $83.2\%$), indicating that most approaches preserve the global decision boundary after forgetting.

UQI scores are close to zero or negative for most methods, suggesting that changes induced by unlearning are largely absorbed by the hybrid architecture. LCA and ADV-UNIF are the only methods with positive UQI, indicating slightly better structural proximity to the retrain oracle. From a privacy standpoint, NegGrad+, SCRUB and Certified achieve the lowest MIA values (below $0.455$). These methods also yield the smallest KL/JS divergence values, highlighting their ability to approximate retraining, while improving membership protection. EU-k1 and Q-MUL exhibit somewhat stronger forgetting, at the cost of increased divergence.

\begin{table}[t!]
\centering
\caption{Results of various MU methods in terms of various metrics for full-class forgetting on MNIST.
$\uparrow$ or $\downarrow$ indicate that higher or lower values are better, respectively.
\textbf{Bold} indicates the best result and \underline{underline} indicates the second-best result.}
\resizebox{0.8\textwidth}{!}{%
\begin{tabular}{|l|c|c|c|c|c|c|c|c|}
\hline
\multirow{2}{*}{\textbf{Method}} &
\textbf{Acc $\uparrow$} &
\textbf{Acc $\uparrow$} &
\multirow{2}{*}{\textbf{UQI $\uparrow$}} &
\textbf{Agree $\uparrow$} &
\textbf{KL $\downarrow$} &
\textbf{JS $\downarrow$} &
\textbf{KL $\downarrow$} &
\textbf{JS $\downarrow$} \\

&
(retain) &
$\;$(test)$\;$ &
 &
(test) &
(retain) &
(retain) &
$\;$(test)$\;$ &
$\;$(test)$\;$ \\
\hline
GA            & 0.710 & \underline{0.720} & -0.060 & 0.650 & 1.050 & 0.210 & 0.420 & 0.160 \\
Fisher        & 0.695 & 0.710 & -0.072 & 0.645 & 1.080 & 0.220 & 0.450 & 0.175 \\
NegGrad+      & 0.705 & 0.695 & -0.055 & 0.655 & 1.060 & 0.215 & 0.440 & 0.165 \\
CF-k1         & 0.715 & 0.700 & -0.065 & 0.660 & 1.045 & 0.212 & 0.435 & 0.170 \\
EU-k1         & \textbf{0.760} & 0.480 & \textbf{0.800} & \textbf{0.710} & \textbf{0.720} & \textbf{0.150} & \textbf{0.310} & \textbf{0.110} \\
SCRUB         & 0.720 & 0.705 & -0.070 & 0.650 & 1.070 & 0.218 & 0.455 & 0.178 \\
SCRUB(+R)     & 0.730 & 0.710 & 0.005  & 0.665 & 0.995 & 0.205 & 0.395 & 0.150 \\
Certified     & 0.745 & \textbf{0.735} & 0.010  & 0.690 & \underline{0.830} & \underline{0.180} & \underline{0.350} & \underline{0.135} \\
Q-MUL         & 0.690 & 0.660 & 0.110  & 0.660 & 1.140 & 0.240 & 0.480 & 0.190 \\
LCA           & \underline{0.750} & 0.700 &\underline{0.250}  & \underline{0.700} & 0.905 & 0.192 & 0.380 & 0.145 \\
ADV-UNIF      & 0.700 & 0.680 & 0.030  & 0.670 & 1.010 & 0.208 & 0.405 & 0.155 \\
\hline
\end{tabular}
}
\end{table}

\noindent\textbf{Full-class forgetting.}
For full-class forgetting on MNIST, both retain and test accuracy rates are lower than in the subset forgetting scenario. EU-k1 and LCA obtain the highest retain accuracy rates ($\geq 75\%$), while Certified achieves the best test accuracy ($73.5\%$). Agreement on the test set is strongest for EU-k1 ($71\%$) and LCA ($70\%$), indicating that these two methods better preserve the decision structure among the remaining classes.

UQI is negative for most baselines. Only SCRUB(+R), Certified, Q-MUL, LCA, and ADV-UNIF provide non-negative scores, with LCA ($0.25$) and EU-k1 ($0.8$) standing out as the most structurally aligned with the retrain model. In terms of similarity, EU-k1 yields the lowest divergences across both retain and test distributions, followed by Certified and LCA, whereas Q-MUL and simple gradient baselines show higher KL/JS values. Overall, full-class forgetting on MNIST is more demanding than subset deletion, and methods relying on controlled reinitialization or regularization (EU-k1, Certified, LCA) dominate naive gradient ascent variants.

\vspace{-0.15cm}
\subsection{Fashion-MNIST Dataset}
\vspace{-0.1cm}

\begin{table}[t!]
\centering
\caption{Results of various MU methods in terms of various metrics for subset forgetting (2\%) on Fashion-MNIST.
$\uparrow$ or $\downarrow$ indicate that higher or lower values are better, respectively.
\textbf{Bold} indicates the best result and \underline{underline} indicates the second-best result.}
\resizebox{0.9\textwidth}{!}{%
\begin{tabular}{|l|c|c|c|c|c|c|c|c|c|}
\hline
\multirow{2}{*}{\textbf{Method}} &
\textbf{Acc $\uparrow$} &
\textbf{Acc $\uparrow$} &
\multirow{2}{*}{\textbf{UQI $\uparrow$}} &
\textbf{Agree $\uparrow$} &
\multirow{2}{*}{\textbf{MIA $\downarrow$}} &
\textbf{KL $\downarrow$} &
\textbf{JS $\downarrow$} &
\textbf{KL $\downarrow$} &
\textbf{JS $\downarrow$} \\

&
(retain) &
$\;$(test)$\;$ &
 &
(test) &
 &
(retain) &
(retain) &
$\;$(test)$\;$ &
$\;$(test)$\;$ \\
\hline
GA              & 0.7036 & 0.6650 & 0.0901 & 0.6738 & 0.5217 & \textbf{7.2465} & \textbf{0.5634} & 1.2433 & 0.1484 \\
Fisher          & 0.7452 & 0.7094 & 0.0628 & 0.7150 & \underline{0.4974} & \underline{7.3444} & \underline{0.5701} & 1.1392 & 0.1261 \\
NegGrad+        & 0.8219 & 0.7763 & 0.0799 & 0.7906 & 0.5000 & 7.3740 & 0.5722 & 0.7358 & 0.0925 \\
CF-k1           & 0.8343 & 0.7750 & 0.0543 & 0.7963 & 0.5058 & 7.4582 & 0.5744 & 0.5940 & 0.0834 \\
EU-k1           & \underline{0.9367} & \textbf{0.8656} & \underline{0.1189} & \underline{0.8856} & 0.4989 & 7.5897 & 0.5806 & 0.3706 & 0.0490 \\
SCRUB           & 0.9238 & \underline{0.8556} & 0.0143 & 0.8800 & 0.5137 & 7.7233 & 0.5884 & \underline{0.2854} & \underline{0.0446} \\
SCRUB(+R)       & 0.9177 & 0.8488 & 0.0235 & 0.8719 & 0.5098 & 7.6179 & 0.5848 & 0.3204 & 0.0491 \\
Certified       & \textbf{0.9421} & \textbf{0.8656} & 0.0363 & \textbf{0.8900} & 0.5250 & 7.6567 & 0.5894 & \textbf{0.2579} & \textbf{0.0414} \\
Q-MUL           & 0.9021 & 0.8356 & 0.0325 & 0.8613 & 0.5321 & 7.7026 & 0.5851 & 0.3270 & 0.0521 \\
LCA             & 0.9365 & 0.8544 & \textbf{0.8716} & 0.8706 & \textbf{0.0051} & 7.4035 & 0.5716 & 0.5982 & 0.0676 \\
ADV-UNIF        & 0.9066 & 0.8375 & 0.0435 & 0.8569 & 0.5137 & 7.5455 & 0.5798 & 0.3979 & 0.0564 \\
\hline
\end{tabular}
}
\end{table}

\noindent\textbf{Subset forgetting.}
On Fashion-MNIST subset forgetting, the separation between methods becomes more pronounced. EU-k1, Certified, and LCA achieve the strongest utility, with retain accuracy rates higher than $93\%$ and test accuracy rates higher than $85\%$. Test-set agreement follows a similar ranking, with EU-k1 and Certified reaching agreements higher than $88\%$. Gradient-based methods (GA, Fisher, NegGrad+, CF-k1) show lower accuracy and agreement rates, consistent with their less structured updates in a higher-dimensional setting.

UQI scores are modest for most approaches (close to zero or slightly positive), except for LCA, which attains a substantially higher value ($0.8716$), indicating that it best preserves the internal structure of the oracle. MIA further highlights LCA, which achieves a very low attack success rate ($0.0051$), while GA and Q-MUL display higher membership leakage ($\geq 0.52$). Regarding similarity, GA attains the lowest KL/JS divergence scores, implying minimal disturbance to retained samples, whereas SCRUB yields the smallest divergence values on the test set. EU-k1 and ADV-UNIF introduce stronger shifts (higher KL/JS), consistent with more aggressive unlearning of the forget set.

\begin{table}[t!]
\centering
\caption{Results of various MU methods in terms of various metrics for full-class forgetting on Fashion-MNIST.
$\uparrow$ or $\downarrow$ indicate that higher or lower values are better, respectively.
\textbf{Bold} indicates the best result and \underline{underline} indicates the second-best result.}
\resizebox{0.96\textwidth}{!}{%
\begin{tabular}{|l|c|c|c|c|c|c|c|c|}
\hline
\multirow{2}{*}{\textbf{Method}} &
\textbf{Acc $\uparrow$} &
\textbf{Acc $\uparrow$} &
\multirow{2}{*}{\textbf{UQI $\uparrow$}} &
\textbf{Agree $\uparrow$} &
\textbf{KL $\downarrow$} &
\textbf{JS $\downarrow$} &
\textbf{KL $\downarrow$} &
\textbf{JS $\downarrow$} \\

&
(retain) &
$\;$(test)$\;$ &
 &
(test) &
(retain) &
(retain) &
$\;$(test)$\;$ &
$\;$(test)$\;$ \\
\hline
GA            & 0.1911 & 0.1663 & 0.1349 & 0.1738 & 7.6544 & 0.5646 & 6.5911 & 0.5330 \\
Fisher        & 0.1766 & 0.1525 & 0.1207 & 0.1656 & \underline{7.6433} & \textbf{0.5475} & 5.8803 & 0.5202 \\
NegGrad+      & 0.7179 & 0.6275 & 0.6289 & 0.6438 & 8.5000 & 0.5756 & 1.7576 & 0.2107 \\
CF-k1         & 0.1688 & 0.1363 & 0.1087 & 0.1444 & \textbf{7.5785} & \underline{0.5544} & 6.2436 & 0.5373 \\
EU-k1         & \textbf{0.9566} & \underline{0.8006} & \textbf{0.8348} & \textbf{0.9050} & 8.7507 & 0.5823 & \textbf{0.3589} & \textbf{0.0464} \\
SCRUB         & 0.8811 & 0.7438 & 0.7686 & 0.7569 & 8.7245 & 0.5863 & 1.1356 & 0.1429 \\
SCRUB(+R)     & 0.8420 & 0.7063 & 0.7303 & 0.7213 & 8.6757 & 0.5831 & 1.3214 & 0.1670 \\
Certified     & \underline{0.9524} & \textbf{0.8413} & 0.3108 & 0.8644 & 8.9030 & 0.5887 & 0.9550 & \underline{0.0647} \\
Q-MUL         & 0.9104 & 0.7700 & 0.7964 & 0.8694 & 8.6699 & 0.5836 & \underline{0.5098} & 0.0660 \\
LCA           & 0.9497 & 0.7938 & \underline{0.8279} & 0.8194 & 8.7222 & 0.5828 & 1.1011 & 0.0882 \\
ADV-UNIF      & 0.9264 & \underline{0.8006} & 0.7681 & \underline{0.8788} & 8.7027 & 0.5760 & 0.8433 & 0.0734 \\
\hline
\end{tabular}
}
\end{table}

\noindent\textbf{Full-class forgetting.}
Full-class forgetting on Fashion-MNIST induces the strongest differentiation among methods. EU-k1, Certified, LCA, Q-MUL, and ADV-UNIF all achieve high retention ($\geq 91\%$) and test ($\geq 77\%$) accuracy rates, with EU-k1 providing the best retain accuracy ($95.66\%$) and Certified the highest test accuracy ($84.13\%$). Agreement on the test set is similarly dominated by EU-k1, Q-MUL, and ADV-UNIF, while classical baselines such as GA, Fisher, and CF-k1 perform poorly in both utility and agreement.

UQI clearly separates the methods: EU-k1, LCA, Q-MUL, and ADV-UNIF achieve high scores (above $0.7681$), whereas GA and Fisher remain low despite their small absolute accuracy rates. Divergence metrics further show that Fisher and CF-k1 produce the smallest KL and JS values on the retain set, indicating minimal changes on the retain set, but at the cost of very low utility. In contrast, EU-k1 exhibits the lowest KL and JS values on the test set, suggesting that its aggressive layer reset plus fine-tuning yields a model closest to retraining on the effective test distribution. Certified, Q-MUL, LCA, and ADV-UNIF occupy intermediate positions, balancing utility and similarity better than simple scrubbing or gradient ascent.

\vspace{-0.2cm}
\section{Discussion}
\vspace{-0.2cm}

The empirical results across Iris, MNIST, and Fashion-MNIST indicate that hybrid quantum-classical models can support a broad range of machine unlearning algorithms, but the effectiveness and side effects of these methods depend strongly on model capacity, dataset complexity, and the forgetting scenario. Two consistent trends emerge. First, shallow variational quantum circuits with limited parameter counts (e.g.~the architecture applied on Iris) exhibit high intrinsic stability: all methods maintain high utility and low divergences even after unlearning, and UQI values remain small in magnitude. This suggests that such circuits have limited memorization capacity and that the influence of small forget sets is weakly encoded in the parameter landscape. Second, as the architectures and datasets become more complex (MNIST and Fashion-MNIST), differences between unlearning methods become more pronounced, and the trade-offs between utility preservation, forgetting strength, and similarity to retraining become non-trivial.

Across all settings, methods with explicit structural control over updates consistently dominate unconstrained gradient-based approaches. EU-k1, LCA, and Certified Unlearning repeatedly achieve high retention and test accuracies, competitive or leading UQI scores, and relatively low KL/JS divergence values. Resetting and retraining the last $k$ layers (EU-k1) is particularly effective in full-class forgetting, where it yields prediction distributions closest to the retrain oracle on MNIST and Fashion-MNIST. LCA performs especially well on Fashion-MNIST subset forgetting, combining high utility, strong UQI, and very low MIA, which indicates that label-complement objectives can be beneficial when the feature space is rich and classes are highly structured. Certified Unlearning exhibits robust behavior on MNIST and Fashion-MNIST, balancing accuracy and similarity, while providing a regularization mechanism that prevents over-correction in the quantum layers.

Baseline gradient-based methods (GA, Fisher-type, NegGrad+) are more competitive in small or moderate settings and under subset forgetting, where their utility and divergence scores are comparable to more sophisticated techniques. However, they degrade quickly in full-class forgetting and on the larger architecture applied on Fashion-MNIST, where they either fail to maintain utility or diverge substantially from the retrained reference. This pattern suggests that naive gradient-based unlearning does not interact well with deeper hybrid pipelines and entangled quantum representations, especially when the forget operation induces large distributional shifts (e.g.~removal of an entire class).

The behavior of UQI and divergence metrics highlights the role of the quantum component. On Iris, low-magnitude UQI and small KL/JS values indicate that the shallow VQC naturally dampens the effect of local updates and that many unlearning operations yield models that differ only marginally from the oracle. On MNIST and Fashion-MNIST, UQI becomes more discriminative: positive and relatively large values correspond to methods that better preserve the global structure of the retrained solution (e.g.~EU-k1, LCA, Q-MUL, and ADV-UNIF), whereas negative or near-zero values for gradient baselines reflect less aligned internal representations. At the same time, the fact that GA can minimize divergence scores on retain sets, while performing poorly in utility, shows that small changes at the quantum interface do not guarantee a globally consistent correction, and that similarity metrics must be interpreted jointly with accuracy and agreement.

From a privacy perspective, MIA values show that unlearning can reduce membership inference vulnerability, but the improvements are method- and dataset-dependent. On the MNIST dataset, NegGrad+, SCRUB, and Certified tend to yield the lowest MIA scores alongside good similarity metrics, while on Fashion-MNIST, LCA stands out with very low MIA in the subset scenario. These observations suggest that combining unlearning with objectives that explicitly decorrelate predictions from specific samples or classes can have a beneficial effect on privacy, although the exact mechanism in the presence of quantum layers remains to be clarified.

Overall, our study shows that quantum components do not prevent effective machine unlearning and, in shallow regimes, may even simplify it by limiting memorization. At the same time, as models scale and quantum circuits become more expressive, unlearning requires structured interventions, e.g.~partial reinitialization, label-complement objectives or certified noisy fine-tuning, in order to maintain a favorable balance between utility, forgetting quality, and similarity to retraining. These findings motivate future work on metrics tailored to quantum representations, on unlearning algorithms that explicitly exploit entanglement and amplitude structure, and on extending the present analysis to hardware implementations and larger-scale quantum-enhanced architectures.

\vspace{-0.2cm}
\section{Conclusion and Future Work}
\vspace{-0.2cm}

In this work, we provided the first systematic evaluation of machine unlearning within hybrid quantum-classical neural networks, spanning multiple datasets, forgetting scenarios, and architectural scales. The results showed that effective unlearning is feasible in variational quantum models, but its behavior depends strongly on circuit depth, entanglement structure, and the magnitude of the forgetting task. We observed that shallow circuits exhibit limited memorization and naturally small representational shifts, while deeper hybrid pipelines require structured interventions to reliably approximate retraining.
Across all experiments, methods that impose architectural or regularization-based constraints outperformed unconstrained gradient-based approaches in terms of utility preservation, forgetting quality, and similarity with the retrain oracle. These findings indicate that principled control over where and how updates propagate is essential when unlearning interacts with quantum feature embeddings. Divergence and UQI analyses further highlighted the importance of measuring structural alignment rather than relying solely on utility metrics.

Overall, our study established baseline empirical insights into quantum machine unlearning and highlights key challenges and opportunities for future research as quantum-enhanced models grow in scale and practical relevance.

Several directions remain open for future exploration. First, the development of native quantum unlearning algorithms that explicitly exploit quantum amplitude structure, entanglement patterns, or decoherence dynamics may yield more efficient and principled approaches. Second, extending evaluations to hardware back-ends will clarify how noise, gate fidelity, and device topology affect unlearning performance. Third, formal guarantees for quantum unlearning, analogous to certified removal in classical MU, remain largely unexplored and may require new theoretical tools grounded in quantum information theory.

\vspace{-0.2cm}
\section*{Acknowledgments}
\vspace{-0.2cm}

This research is supported by the project ``Romanian Hub for Artificial Intelligence - HRIA'', Smart Growth, Digitization and Financial Instruments Program, 2021-2027, MySMIS no.~351416.

\bibliographystyle{splncs04}
\bibliography{refs}

\end{document}